%% file: elsarticle-template-1-num.tex
\documentclass[preprint,12pt]{elsarticle}

\usepackage{graphicx}
\usepackage{amssymb}

\usepackage{lineno}

\usepackage{multirow}

\journal{Journal Name}

\makeatletter
\def\ps@pprintTitle{%
   \let\@oddhead\@empty
   \let\@evenhead\@empty
   \let\@oddfoot\@empty
   \let\@evenfoot\@oddfoot
}
\makeatother

\begin{document}

\begin{frontmatter}


\title{2D/3D Megavoltage Image Registration Using Convolutional Neural Networks}



\author[1]{Hector N. B. Pinheiro}
\author[1]{Tsang Ing Ren}
\author[2]{Stefan Scheib} 
\author[2]{Armel Rosselet}
\author[2]{Stefan Thieme-Marti}

\address[1]{Centro de Inform\'{a}tica-CIn, Universidade Federal de Pernambuco-UFPE, Recife, Brasil}

\address[2]{Varian Medical Systems, Imaging Laboratory GmbH, Baden, Switzerland}

\begin{abstract}
We presented a 2D/3D MV image registration method based on a Convolutional Neural Network.  Most of the traditional image registration method intensity-based, which use optimization algorithms to maximize the similarity between to images.  Although these methods can achieve good results for kilovoltage images, the same does not occur for megavoltage images due to the lower image quality. Also, these methods most of the times do not present a good capture range.  To deal with this problem,  we propose the use of  Convolutional Neural Network.  The experiments were performed using a dataset of 50 brain images. The results showed to be promising compared to traditional image registration methods.

\end{abstract}

\begin{keyword}
2D/3D Image Registration \sep Megavoltage Image  \sep Convolutional Neural Networks


\end{keyword}

\end{frontmatter}


\input{introduction}

\input{proposed}

\input{experiments}

\input{conclusion}

\section*{Acknowledgment}
We would like to thank Helen Khoury, Silvio de Barros Melo, Halisson Alberdan Cavalcanti Cardoso for valuable discussion. Also, Dr. Ernesto Roesler and his team from Hospital Portugu\^es, Lucas Delbem, Karen Pieri, Thiago Fontana for making the CT images available for this research.  






\bibliographystyle{model1-num-names}
\bibliography{sample.bib}







\end{document}

%% file: introduction.tex
\section{Introduction}
\label{sec_introduction}

The correct patient positioning is an essential procedure for the success of radiotherapy treatment. During the preoperative phase, a three-dimensional CT (Computed Tomography) of the planning is used both to set the location of the tumor and to define the treatment parameters (number of sessions, dose, etc.). In subsequent treatment sessions, the patient should be well positioned so that the doses can be distributed at the planned sites. Several techniques are used for guiding the patient position, such as body markings and laser systems, but still positioning errors are a reality. In fact, such errors can reach 10-20 mm in some types of cancer \cite{ramsey, SCHUBERT20091260}. One of the most effective alternatives to reduce these errors consists of Image-Guided Radiation Therapy (IGPR), in which images are acquired and compared with the 3D image of the planning CT to verify the patient's positioning before the application of the treatment dose. The comparison between the images has the objective of estimating the displacements necessary for the correction of the patient's positioning. This task is called Image Registration (IR), which can be performed by visually inspecting the overlapping images (manual registration) or through the use of algorithms that search for the best match between the images (automatic registration). Depending on the machine used, registration can be performed using two-dimensional MegaVoltage (MV) or KiloVoltage (KV) images, or using three-dimensional (\emph{Cone-beam CTs}) images. Typically, the same machine on which the treatment is performed is capable of producing megavoltage images without any additional apparatus, which makes such an approach still the most widely used. In this approach, a pair of two-dimensional MV images is compared to two-dimensional Digitally Reconstructed Radiographs (DRRs) extracted from the planning CT.

The most used methods for 2D/3D registration are intensity-based, which use optimization algorithms to maximize the similarity between the 2D images and the DRRs. Although these methods can achieve good results for kilovoltage images, the same does not occur for megavoltage images due to the lower image quality. Furthermore, such methods do not have a good capture range, which implies the need for numerous maximization iterations. Since at each iteration a 2D DRR is extracted from the CT, such methods are hardly capable of being executed in real time.

To deal with this problem, recently Miao \emph{et. al} proposed the use of a hierarchical regression method for real-time X-ray 2D/3D image registration \cite{miao2016cnn}. The regression is performed applying a Convolutional Neural Network, whose training is performed using two-dimensional artificial X-ray images. The trained model is then able to infer the displacements necessary for the correction of positioning errors without the explicit definition of any measure of similarity.  As the problem of inferring the displacements has extremely high complexity, the authors proposed the use of different regressors, which are specialized in different ranges of displacement values. This hierarchical approach, together with iterative execution, proved to be quite effective, presenting results superior to those achieved by methods based on image similarity. Moreover, such results were achieved with few regression iterations (less than ten), which makes it possible to use the real-time approach.
In this work, we follow the same approach proposed by Miao \emph{et. al}. However, we focused on image registration using MV images. In addition, the regression models have significant differences in their architectures which are described in the next sections.

%% file: proposed.tex
\section{Proposed 2D/3D Registration Method}
\label{sec_regression}

\begin{figure}[ht!]
  \centering
  \includegraphics[width=0.7\textwidth,angle=-90,origin=c]{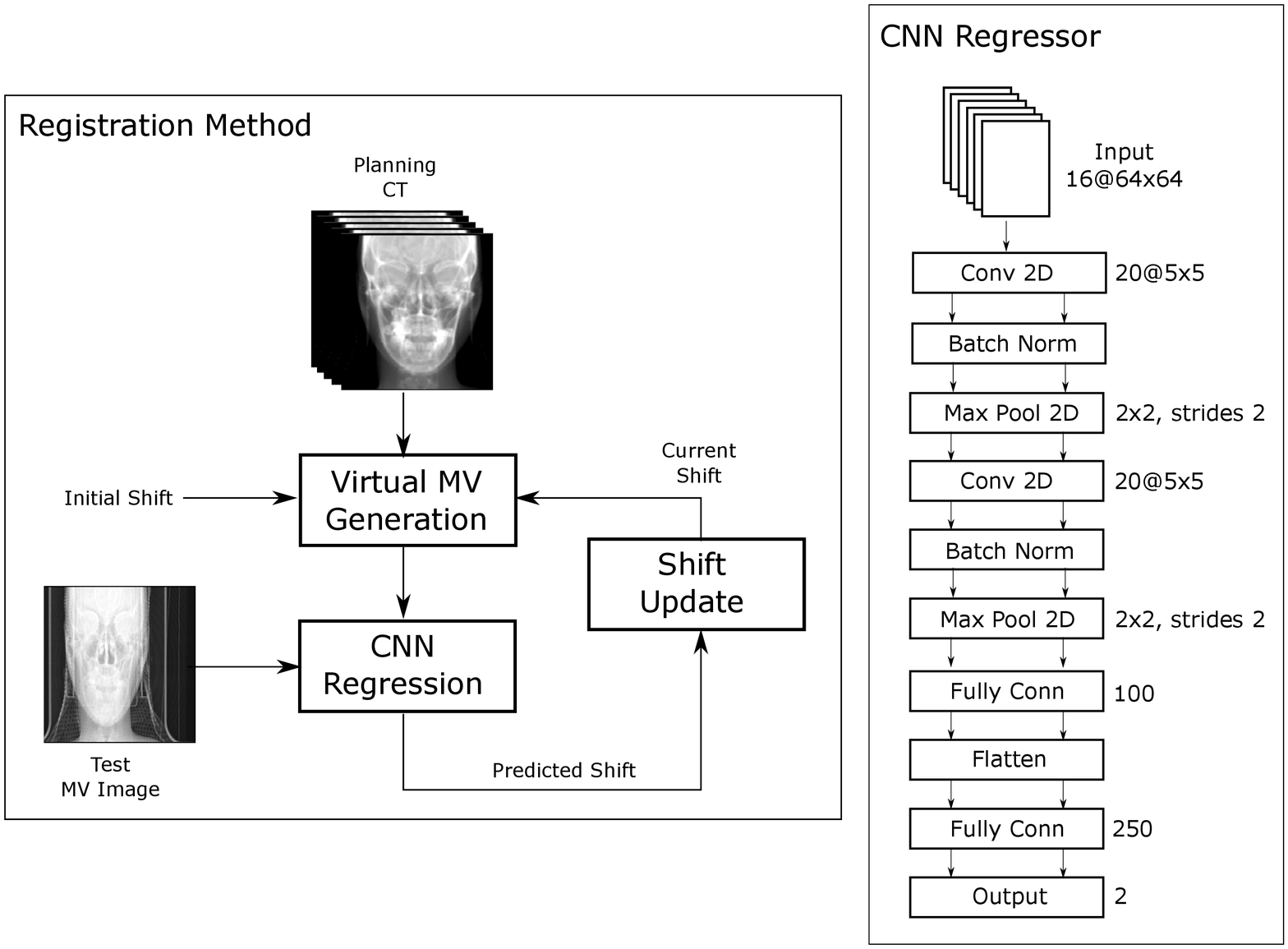}
  \caption{On the right, the scheme of the proposed method, based on regression. Iteratively, the displacement values define the inputs of the regression model, which infers the new displacement values. On the left, the architecture of the CNN network used for the regression.}
  \label{fig_proposed}
\end{figure}

The proposed method is based on the concept of regression. Unlike the similarity-based methods, where an optimization algorithm is employed to find displacements that maximize a given measure of similarity, the proposed method uses a regression model whose parameters are learned through machine learning techniques. Figure \ref{fig_proposed} shows the diagram of this approach. Starting from an initial offset value, the method uses the planning CT to generate an artificial MV image employing the current displacement values. This MV image is then processed by the regression method in conjunction with the test MV image, defining the displacement required for the shifted image to conform to the test image. At each iteration, the result of the regression is used to define the system response.  Given the current displacement values, $\Delta_i$, and the predicted value by the regression model, $\Delta_k$, the new displacement - the response of the registration method to the iteration - is defined by the accumulation of these values: $\Delta_{i+1} = \Delta_i+\Delta_k$. Similar to the traditional optimization methods, this process can be performed iteratively until a certain convergence condition is reached ($\Delta_k \approx 0 $).

The regression model consists of a Convolutional Neural Network, whose parameters are learned with the purpose of estimating the displacement values ​​necessary for two-dimensional images to be coupled (\emph{match}). Through this approach, it is not necessary to fix a measure of similarity to be maximized. From training data, which define known displacement values ​​for image examples, learning algorithms adjust the weights of the network in order to define a model that can be used for arbitrary images. Figure \ref{fig_proposed} shows the proposed network settings. Given the test image and the artificial MV image produced with the current displacement values, we first compute the difference between the images, extracting from them the Region Of Interest (ROI) defined for the processing. The resulting image is then divided into a 4x4 grid, producing a total of 16 sub-images (or \emph{patches}). The network input is defined by a three-dimensional image with 16 channels, one for each \emph{patch}.

The input is then processed by two convolutional processes, which are defined by a convolutional layer with 20 nodes with 5x5 bi-dimensional filters, followed by a normalization layer of \emph{batch} and one of \emph{pooling} \emph{Pool} 2x2 with \emph{strides} = 2). Next, the network is composed of a dense layer (100\% nested), maintaining the volume of the data, which is then flattened and processed by a dense layer with 250 nodes that connect to the output layer which define the two predicted displacement values. Here, we use the ReLU (Rectified Linear Unit) activation function for the convoluted and dense layers.

Using an image database with controlled shifts, the desired outputs for the data presented to the network are used to adjust the layer parameters. This adjustment was performed using the Stochastic Gradient Descending (SGD) algorithm with cost function defined by the Mean Square Error (MSE). The algorithm was executed with a learning rate of 0.0001 (with a decay of the same value), and the value for \emph{momentum} set to 0.9. Initially, the parameters are initialized using the Glorot \cite{glorot2010understanding} method.

%% file: experiments.tex
\section{Experiments}

\subsection{Dataset}

\begin{figure}[ht!]
  \centering
  \includegraphics[width=0.6\textwidth,angle=-90,origin=c]{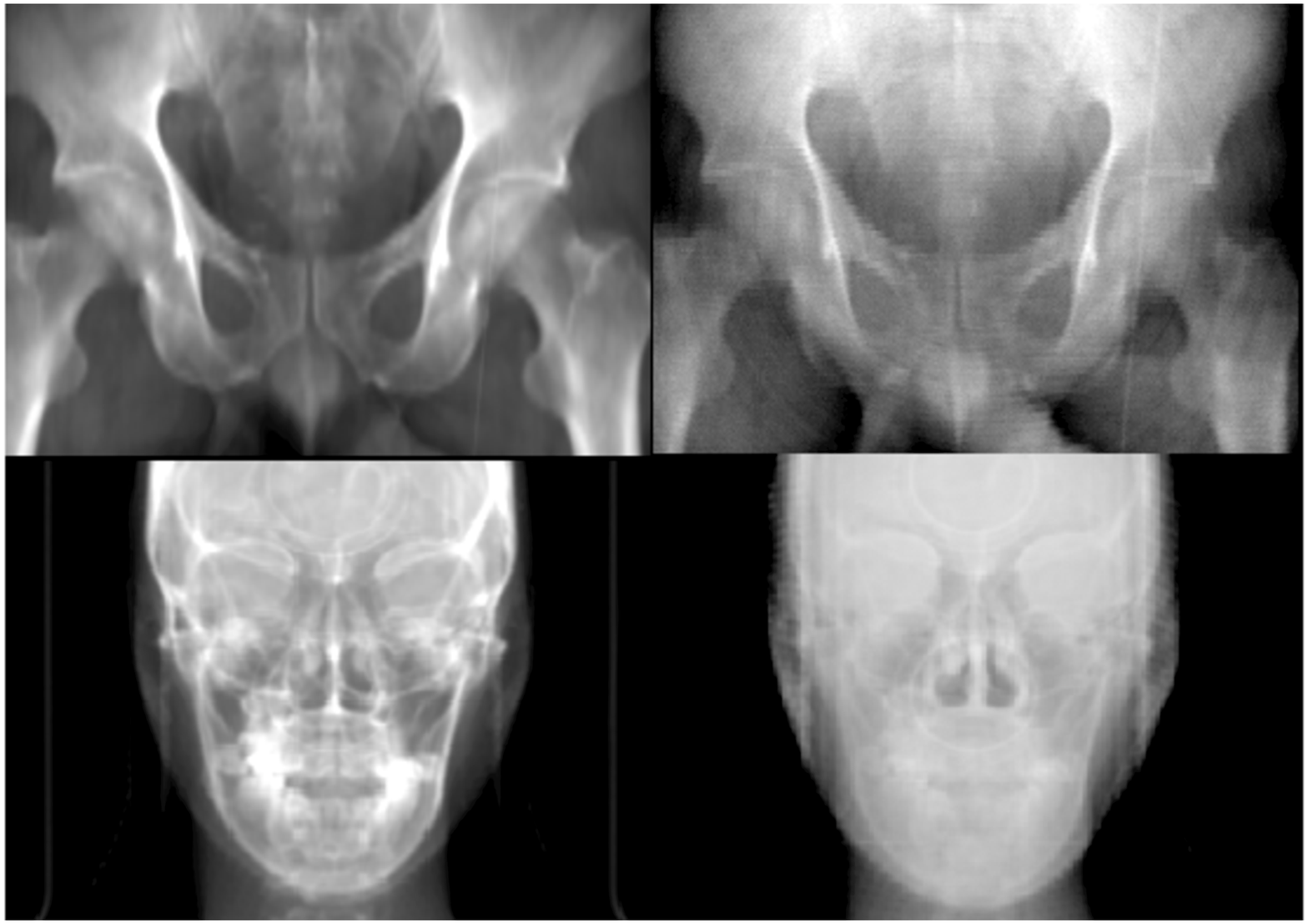}
  \caption{Examples of artificial MV images (right) generated from CT images from the brain and pelvis. The images on the left show the respective DRRs extracted in the same direction that generates the MV images.}
  \label{fig_mv}
\end{figure}

In order to evaluate the proposed method in a scenario close to reality, artificial MV images generated from real CTs were used. Such images were generated following the method proposed by Kieselmann \emph{et. al} \cite{ Kieselmann}, where the Hounsfield Units (HUs) present in the CT are mapped taking into account the differences present in the attenuation values for energies of 80keV (reference from CT ) to 0.3 MeV (reference from the MV images). Also, the model also describes beam scattering and noise addition, present in MV images. Figure \ref{fig_mv} shows examples of DRRs extracted from CTs and artificial MV images generated by the model for brain and pelvic CTs.

\begin{figure}[ht!]
  \centering
  \includegraphics[width=0.6\textwidth]{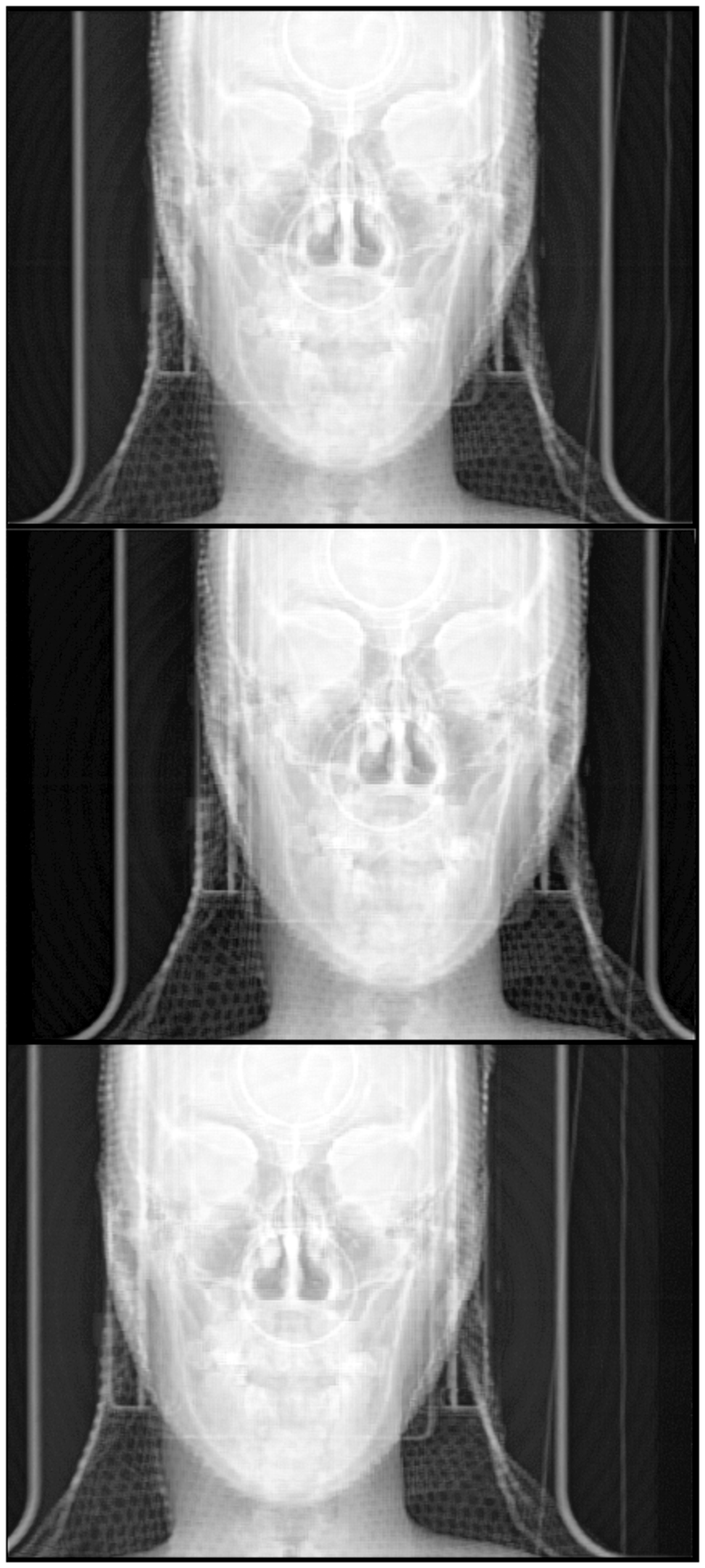}
  \caption{Examples of artificial MV images generated with different longitudinal displacements. While the top image shows no displacement, the middle and the bottom images have displacements of + 2cm and -2cm, respectively.}
  \label{fig_shift}
\end{figure}

In total, 50 images of brain CTs were used for the experiments,  the training set of the regression model was composed of 25 images, and the other 25 were used for evaluation. The test set was produced by fixing the displacement values ​​in the lateral and longitudinal directions, with the displacement values ​​of $\pm$5mm, $\pm$10mm and $\pm$20mm. A total of 6 possible displacements in each direction, generating 36 samples for each test CT.  This configuration produced a total of 900 test registration cases. The training set was produced by randomly generating displacements in both directions, following a uniform distribution [-20,20] mm. For each CT, 100 displacements were created by fixing a direction with zero displacement, and 300 displacements were generated with non-zero values. For each CT, 500 training images were generated, giving a total of 12500 samples. Examples of these images are shown in Figure \ref{fig_shift}.

\subsection{Comparison with State-of-Art Methods}
The proposed method was compared with methods based on similarity which are more widely used.  Each of these methods can be described by the applied optimization algorithm and the similarity measure to be maximized. The optimization occurs taking into account an initial solution. At each iteration, new solutions are generated until the value to be optimized converges, that is, until the difference in the value of the similarity of the solution found does not exceed a certain tolerance threshold $\phi $. In this work, we consider two optimization algorithms: the Powell \cite {van2011evaluation} method and the Downhill Simplex method of Nelder-Mead \cite{rivest2012nonrigid}, and three measures of similarity: Mutual Information (MI) \cite{zollei20012d} , Cross-Correlation (CC) \cite{knaan2003effective} and Pattern Intensity (PI) \cite{russakoff2003evaluation}. For both, the six combinations of intensity-based methods and the proposed method, the initial solution was fixed at the origin, and the Region Of Interest (ROI) was defined as the 10 cm side center square. Also, for methods based on similarity, a tolerance value of $\phi = 0.0001 $ was used.

\subsection{Evaluation Metrics}
The gold standards are defined by the fixed displacements generated for the test set. We compute the distances between the values of displacements generated by the methods and the expected values. Also, based on the evaluation of the margins of error applied in treatments \cite{suzuki2012uncertainty, poulsen2007residual}, we set a cut-off threshold of 2 mm for the distance between the displacement and the gold standard. Then, we define that when such distance is above the threshold, a false positive case occurs. The then false positive rate is defined by the ratio of the number of false positives by the number of test records. Here, both the false positive rate and the distribution of distances values were used to compare the methods.

\subsection{Results}
 
Table \ref{tab_results} shows the obtained results.  Both, the mean distance between the displacements obtained by the methods and the False Positive Rate (FPR) are the shown in this Table. In addition, the distributions of the distances between the displacements and the gold standard are described by the percentiles 10, 25, 50, 75 and 90. The results by the methods based on similarity were obtained by the optimization methods applied until the tolerance threshold  $ \ phi = 0.0001 $ is reached. For the proposed method, the results until the third iteration are presented. The best results for both approaches are described in bold.

\begin{table}[]
\centering
\caption{
Results obtained by the proposed method (CNN) and by methods based on similarity with Powell and Downhill Simplex optimization algorithms and similarity measures, Pattern Intensity (PI), Mutual Information (MI) and Cross Correlation (CC).
Deviations and False Positive Rates (FPRs) are presented in mm and in \%, respectively. For the CNN method just 3 iterations (iter). 
The best results for each approach are described in bold.}
\begin{tabular}{|l|c|c|c|c|c|c|c|}
\hline
\multicolumn{1}{|c|}{\multirow{2}{*}{\textbf{Method}}} & \multicolumn{6}{c|}{\textbf{Shift Deviation Percentile (mm)}}   & \textbf{FPR} \\ \cline{2-7}
\multicolumn{1}{|c|}{}                                 & \textbf{10th} & \textbf{25th} & \textbf{50th} & \textbf{75th} & \textbf{90th} & \textbf{Mean}  &      \textbf{(\%)}                         \\ \hline
Simplex-PI                                             &   0.00        &    0.21       &    0.33       &    0.53       &    25.00       &    5.22             &    16.22                       \\ \hline
Simplex-MI                                             &   0.38        &    0.64       &    1.12       &    2.40       &    21.41       &    5.28             &    27.78                       \\ \hline
\textbf{Simplex-CC}                                    &   \textbf{0.34}  & \textbf{0.49} &    \textbf{0.68}       &    \textbf{0.91}       &    \textbf{1.08}        &    \textbf{0.70}             &    \textbf{0.00}                       \\ \hline
Powell-PI                                              &   0.19        &    0.40       &    18.21      &    23.83      &    31.00       &    139.10           &    66.89                       \\ \hline
Powell-MI                                              &   0.34        &    0.66       &    1.41       &    20.18      &    27.53       &    10.66            &    44.56                       \\ \hline
Powell-CC                                              &   0.22        &    0.38       &    0.69       &    18.45      &    30.89       &    8.78             &    31.56                       \\ \hline \hline
CNN - 1 iter                                      &   0.46        &    0.77       &    1.14       &    3.08       &    20.90       &    6.37             &    37.22                       \\ \hline
CNN - 2 iters                                     &   0.36        &    0.59       &    0.89       &    2.37       &    16.22       &    4.93             &    26.81                       \\ \hline
\textbf{CNN - 3 iters}                                     &   \textbf{0.23}        &    \textbf{0.38}       &    \textbf{0.55}       &    \textbf{1.48}       &    \textbf{10.29}       &    \textbf{4.15}             &    \textbf{12.79}                       \\ \hline

\end{tabular}
\label{tab_results}
\end{table}

Among the optimization methods, it was observed that the best results were obtained with the Downhill Simplex method. The Powell method proved unstable, where high deviations were observed for some examples. The worst results were observed with the use of PI similarity, where more than half of the registration presented deviations close to 2 cm. No measure of similarity presented results superior to no results achieved by the Simplex method.

Among similarity measures, the best results were observed with CC similarity. In both optimization methods, a better accuracy was observed using this similarity. By combining the Simplex method with CC similarity, we obtained the best results, arriving at a perfect false positive rate. That is, in none of the tests did the deviation observed by the obtained displacement exceed 2 mm - the mean deviation does not reach 1mm. The second best result achieved by the similarity methods was the Simplex-PI, with a false positive rate of 16.22\%. The best performance achieved by the Powell optimization method showed a false positive rate of 31.56\% (Powell-CC).

The proposed method presented a clear performance gain at each new iteration performed. With a single iteration, the FPR rate of 37.22\% was obtained.  In addition, at least half of the test cases showed deviations of at most 1.14mm, showing a good performance achieved by the training of the regression model. In the next iteration, a performance gain of 27.91\% was observed, presenting an FPR rate of 26.81\%, which indicates that its competitive with the Simplex-MI method. Already in the third iteration, a performance gain of 52.29\% was observed, with an FPR  rate of 12.79\%. In this case, more than 75\% of the test cases showed a deviation of less than 2 mm. With an average deviation of 4.15 mm, this result was better than the Simplex-PI method. In this way, it is possible to observe that the proposed method can achieve competitive results with similarity-based methods with very few iterations. Comparing with these methods, depending on the adopted tolerance value, it can reach dozens of iterations, having a method capable of obtaining similar results with fewer iterations is quite attractive, especially from the point of view of speed.

%% file: conclusion.tex
\section{Conclusions}
In the paper, we presented a 2d/3D MV image registration method based on a Convolutional Neural Network. The experiments were performed using a dataset of 50 images from the brain and the results were compared to traditional image registration methods. For the experiments using the CNN, due to computational limitations just 3 interactions of the proposed model were performed. However, the results show to be promising since for each interaction the FPR show a decrease of approximately 30\%. Further investigation is needed to evaluate the full potential of the method. 

%% file: elsarticle-template-1-num.bbl
\begin{thebibliography}{12}
\expandafter\ifx\csname natexlab\endcsname\relax\def\natexlab#1{#1}\fi
\providecommand{\bibinfo}[2]{#2}
\ifx\xfnm\relax \def\xfnm[#1]{\unskip,\space#1}\fi
\bibitem[{Ramsey et~al.(2007)Ramsey, Scaperoth, Seibert, Chase, Byrne, and
  Mahan}]{ramsey}
\bibinfo{author}{C.~R. Ramsey}, \bibinfo{author}{D.~Scaperoth},
  \bibinfo{author}{R.~Seibert}, \bibinfo{author}{D.~Chase},
  \bibinfo{author}{T.~Byrne}, \bibinfo{author}{S.~Mahan},
\newblock \bibinfo{title}{Image-guided helical tomotherapy for localized
  prostate cancer: technique and initial clinical observations},
\newblock \bibinfo{journal}{Journal of applied clinical medical physics}
  \bibinfo{volume}{8} (\bibinfo{year}{2007}) \bibinfo{pages}{37--51}.
\bibitem[{Schubert et~al.(2009)Schubert, Westerly, Tomé, Mehta, Soisson,
  Mackie, Ritter, Khuntia, Harari, and Paliwal}]{SCHUBERT20091260}
\bibinfo{author}{L.~K. Schubert}, \bibinfo{author}{D.~C. Westerly},
  \bibinfo{author}{W.~A. Tomé}, \bibinfo{author}{M.~P. Mehta},
  \bibinfo{author}{E.~T. Soisson}, \bibinfo{author}{T.~R. Mackie},
  \bibinfo{author}{M.~A. Ritter}, \bibinfo{author}{D.~Khuntia},
  \bibinfo{author}{P.~M. Harari}, \bibinfo{author}{B.~R. Paliwal},
\newblock \bibinfo{title}{A comprehensive assessment by tumor site of patient
  setup using daily mvct imaging from more than 3,800 helical tomotherapy
  treatments},
\newblock \bibinfo{journal}{International Journal of Radiation
  Oncology/Biology/Physics} \bibinfo{volume}{73} (\bibinfo{year}{2009})
  \bibinfo{pages}{1260 -- 1269}.
\bibitem[{Miao et~al.(2016)Miao, Wang, and Liao}]{miao2016cnn}
\bibinfo{author}{S.~Miao}, \bibinfo{author}{Z.~J. Wang},
  \bibinfo{author}{R.~Liao},
\newblock \bibinfo{title}{A {CNN} regression approach for real-time 2d/3d
  registration},
\newblock \bibinfo{journal}{{IEEE} Transactions on medical imaging}
  \bibinfo{volume}{35} (\bibinfo{year}{2016}) \bibinfo{pages}{1352--1363}.
\bibitem[{Glorot and Bengio(2010)}]{glorot2010understanding}
\bibinfo{author}{X.~Glorot}, \bibinfo{author}{Y.~Bengio},
\newblock \bibinfo{title}{Understanding the difficulty of training deep
  feedforward neural networks},
\newblock in: \bibinfo{booktitle}{Proceedings of the thirteenth international
  conference on artificial intelligence and statistics}, pp.
  \bibinfo{pages}{249--256}.
\bibitem[{Kieselmann et~al.(2016)Kieselmann, Rosselet, Scheib, and
  Thieme-Marti}]{Kieselmann}
\bibinfo{author}{J.~Kieselmann}, \bibinfo{author}{A.~Rosselet},
  \bibinfo{author}{S.~Scheib}, \bibinfo{author}{S.~Thieme-Marti},
\newblock \bibinfo{title}{A systematic analysis of rigid image registration
  using patient cts and simulated setup images with a unique gold standard
  registration},
\newblock \bibinfo{journal}{Medical Physics} \bibinfo{volume}{42}
  (\bibinfo{year}{2016}) \bibinfo{pages}{3291--3292}.
\bibitem[{Van~der Bom et~al.(2011)Van~der Bom, Klein, Staring, Homan, Bartels,
  and Pluim}]{van2011evaluation}
\bibinfo{author}{I.~Van~der Bom}, \bibinfo{author}{S.~Klein},
  \bibinfo{author}{M.~Staring}, \bibinfo{author}{R.~Homan},
  \bibinfo{author}{L.~W. Bartels}, \bibinfo{author}{J.~P. Pluim},
\newblock \bibinfo{title}{Evaluation of optimization methods for
  intensity-based {2D-3D} registration in {X}-ray guided interventions},
\newblock in: \bibinfo{booktitle}{Medical Imaging 2011: Image Processing},
  volume \bibinfo{volume}{7962}, \bibinfo{organization}{International Society
  for Optics and Photonics}, p. \bibinfo{pages}{796223}.
\bibitem[{Rivest-Henault et~al.(2012)Rivest-Henault, Sundar, and
  Cheriet}]{rivest2012nonrigid}
\bibinfo{author}{D.~Rivest-Henault}, \bibinfo{author}{H.~Sundar},
  \bibinfo{author}{M.~Cheriet},
\newblock \bibinfo{title}{Nonrigid {2D/3D} registration of coronary artery
  models with live fluoroscopy for guidance of cardiac interventions},
\newblock \bibinfo{journal}{IEEE Transactions on Medical Imaging}
  \bibinfo{volume}{31} (\bibinfo{year}{2012}) \bibinfo{pages}{1557--1572}.
\bibitem[{Zollei et~al.(2001)Zollei, Grimson, Norbash, and
  Wells}]{zollei20012d}
\bibinfo{author}{L.~Zollei}, \bibinfo{author}{E.~Grimson},
  \bibinfo{author}{A.~Norbash}, \bibinfo{author}{W.~Wells},
\newblock \bibinfo{title}{{2D-3D} rigid registration of x-ray fluoroscopy and
  ct images using mutual information and sparsely sampled histogram
  estimators},
\newblock in: \bibinfo{booktitle}{Proceedings of the 2001 IEEE Computer Society
  Conference o Computer Vision and Pattern Recognition, 2001},
  volume~\bibinfo{volume}{2}, \bibinfo{organization}{IEEE}, pp.
  \bibinfo{pages}{II--II}.
\bibitem[{Knaan and Joskowicz(2003)}]{knaan2003effective}
\bibinfo{author}{D.~Knaan}, \bibinfo{author}{L.~Joskowicz},
\newblock \bibinfo{title}{Effective intensity-based {2D/3D} rigid registration
  between fluoroscopic x-ray and ct},
\newblock in: \bibinfo{booktitle}{International Conference on Medical Image
  Computing and Computer-Assisted Intervention},
  \bibinfo{organization}{Springer}, pp. \bibinfo{pages}{351--358}.
\bibitem[{Russakoff et~al.(2003)Russakoff, Rohlfing, Ho, Kim, Shahidi, Adler,
  and Maurer}]{russakoff2003evaluation}
\bibinfo{author}{D.~B. Russakoff}, \bibinfo{author}{T.~Rohlfing},
  \bibinfo{author}{A.~Ho}, \bibinfo{author}{D.~H. Kim},
  \bibinfo{author}{R.~Shahidi}, \bibinfo{author}{J.~R. Adler},
  \bibinfo{author}{C.~R. Maurer},
\newblock \bibinfo{title}{Evaluation of intensity-based {2D-3D} spine image
  registration using clinical gold-standard data},
\newblock in: \bibinfo{booktitle}{International Workshop on Biomedical Image
  Registration}, \bibinfo{organization}{Springer}, pp.
  \bibinfo{pages}{151--160}.
\bibitem[{Suzuki et~al.(2012)Suzuki, Tateoka, Shima, Yaegashi, Fujimoto,
  Saitoh, Nakata, Abe, Nakazawa, Sakata et~al.}]{suzuki2012uncertainty}
\bibinfo{author}{J.~Suzuki}, \bibinfo{author}{K.~Tateoka},
  \bibinfo{author}{K.~Shima}, \bibinfo{author}{Y.~Yaegashi},
  \bibinfo{author}{K.~Fujimoto}, \bibinfo{author}{Y.~Saitoh},
  \bibinfo{author}{A.~Nakata}, \bibinfo{author}{T.~Abe},
  \bibinfo{author}{T.~Nakazawa}, \bibinfo{author}{K.~Sakata}, et~al.,
\newblock \bibinfo{title}{Uncertainty in patient set-up margin analysis in
  radiation therapy},
\newblock \bibinfo{journal}{Journal of radiation research} \bibinfo{volume}{53}
  (\bibinfo{year}{2012}) \bibinfo{pages}{615--619}.
\bibitem[{Poulsen et~al.(2007)Poulsen, Muren, and
  H{\o}yer}]{poulsen2007residual}
\bibinfo{author}{P.~R. Poulsen}, \bibinfo{author}{L.~P. Muren},
  \bibinfo{author}{M.~H{\o}yer},
\newblock \bibinfo{title}{Residual set-up errors and margins in on-line
  image-guided prostate localization in radiotherapy},
\newblock \bibinfo{journal}{Radiotherapy and Oncology} \bibinfo{volume}{85}
  (\bibinfo{year}{2007}) \bibinfo{pages}{201--206}.

\end{thebibliography}
